\documentclass[10pt,twocolumn,letterpaper]{article}

\usepackage{iccv}
\usepackage{times}
\usepackage{epsfig}
\usepackage{graphicx}
\usepackage{amsmath}
\usepackage{amssymb}
\usepackage{algorithm2e}
\usepackage{authblk}
\usepackage{enumitem}
\setlist{nolistsep}

\usepackage[pagebackref=true,breaklinks=true,letterpaper=true,colorlinks,bookmarks=false]{hyperref}

\iccvfinalcopy 

\def\iccvPaperID{CVAMD-61} 

\ificcvfinal\fi

\newcommand\subfigure[1]{#1}

\begin{document}

\title{Towards Fixing Clever-Hans Predictors with Counterfactual Knowledge Distillation}

\author{
    Sidney Bender$^{1}$ \quad
Christopher J. Anders$^{1,2}$ \quad
Pattarawat Chormai$^{1,3}$ \quad Heike Marxfeld$^4$ \quad \mbox{Jan Herrmann$^4$} \quad Gr\'egoire Montavon$^{5,2,}$\thanks{\textit{Corresponding author:} \texttt{gregoire.montavon@fu-berlin.de}} \\

    \textsuperscript{1}{Machine Learning Group, Technische Universit\"at Berlin, Germany} \\
    \textsuperscript{2}{BIFOLD -- Berlin Institute for the Foundations of Learning and Data, Berlin, Germany} \\
    \textsuperscript{3}{Max Planck School of Cognition, Leipzig, Germany} \\
    \textsuperscript{4}{BASF SE, Ludwigshafen am Rhein, Germany} \\
    \textsuperscript{5}{Department of Mathematics and Computer Science, Freie Universit\"at Berlin, Germany}
}




\maketitle
\ificcvfinal\thispagestyle{empty}\fi

\begin{abstract}

This paper introduces a novel technique called counterfactual knowledge distillation (CFKD) to detect and remove reliance on confounders in deep learning models with the help of human expert feedback. Confounders are spurious features that models tend to rely on, which can result in unexpected errors in regulated or safety-critical domains. The paper highlights the benefit of CFKD in such domains and shows some advantages of counterfactual explanations over other types of explanations. We propose an experiment scheme to quantitatively evaluate the success of CFKD and different teachers that can give feedback to the model. We also introduce a new metric that is better correlated with true test performance than validation accuracy. The paper demonstrates the effectiveness of CFKD on synthetically augmented datasets and on real-world histopathological datasets.

\end{abstract}

\section{Introduction}

Deep learning has made remarkable progress in recent years. Despite this progress, models are still not well understood and often rely on spurious features, i.e. \emph{confounders}, leading to so-called Clever Hans predictors \cite{pfungst1911clever,lapuschkin2019unmasking}. Such predictors may fit the training set well and even maximize validation accuracy, but may result in fatal errors when used in practice. This is particularly important in regulated and safety-critical domains such as medical image processing or autonomous driving, where avoiding such failures is crucial.

In this paper, we introduce a novel technique called counterfactual knowledge distillation (CFKD) to \emph{detect} and \emph{remove} reliance on confounders with the help of human expert feedback. The idea consists of generating counterfactuals that inform the human-expert on the model's decision strategy and leveraging the human-expert's feedback to arrive at a model that is more accurate and is less reliant on confounders. Our contributions are as follows:
\begin{itemize}
    \item We motivate the practical need for CFKD, and present a use case for ovarian follicle detection from histopathology data. 
    \item We highlight some advantages of counterfactual explanations over common attribution-based explanations for \emph{finding} confounders
    \item We develop a methodology based on augmented or resampled datasets for quantitatively evaluating the success our method at \emph{reducing} Clever Hans effects
    \item We demonstrate how CFKD can be used with different forms of explanatory feedback such as \emph{human-in-the-loop} or \emph{oracle models}. 
    \item We propose a new metric which we show empirically to be better correlated with the true test accuracy than the usual accuracy on the validation set.
\end{itemize}
We show that our method works quantitatively on 1) a sanity check dataset, 2) a family of  artificial datasets that check for different types of confounders and different strength of correlation between the confounder and the true feature and 3) a histopathological dataset with a known confounder.

\section{Motivating Example}
\label{section:motivation}

\begin{figure}[h]
  \centering
  \includegraphics[width=0.3\paperwidth]{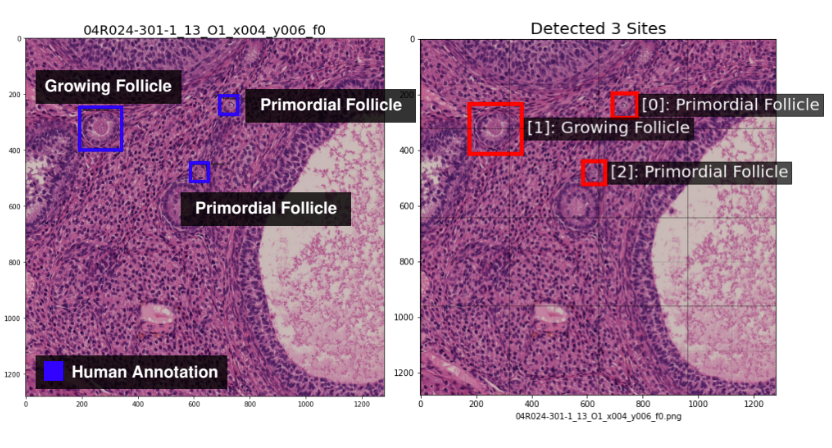}
  \caption{
    \textit{Left:} follicle annotations provided by a human expert on a histopathological slide. \textit{Right:} predictions by the Yolo model.
  }
  \label{fig:yolo_on_follicle_dataset}
\end{figure}

The original motivation for our work lies in the application of machine learning techniques to detect, classify and quantify ovarian follicles in histopathology images\footnote{as required for the assessment of reproductive toxicity according to OECD guidelines 443 and 416. More specifically, primordial follicles and growing follicles (with a granulosa cell layer of up to 5 cells of thickness) in the ovary have to be quantified.}. The tissue samples are cut into slices and the so-called growing and primordial follicles can be counted (see Figure ~\ref{fig:yolo_on_follicle_dataset}).

Manual counting and classification of follicles is a tedious and time-consuming task that requires highly-trained personnel. 
Therefore, these analyses would benefit greatly from computerized automation. 

Recently, architectures like Yolo~\cite{redmon2016you} have shown to be very successful at detecting visual objects in a broad range of image data, including histopathology \cite{sheikhzadeh2018automatic}.
Currently, EU legislation for AI systems is being prepared. Despite the fact that GLP validation is not in the scope of such legislation, the prospective requirements for AI systems in a regulated environment like in toxicology 
give reasons for a comprehensive investigation of the explainability of models in such systems. 

\begin{figure}[h]
  \centering
  \includegraphics[width=0.3\paperwidth]{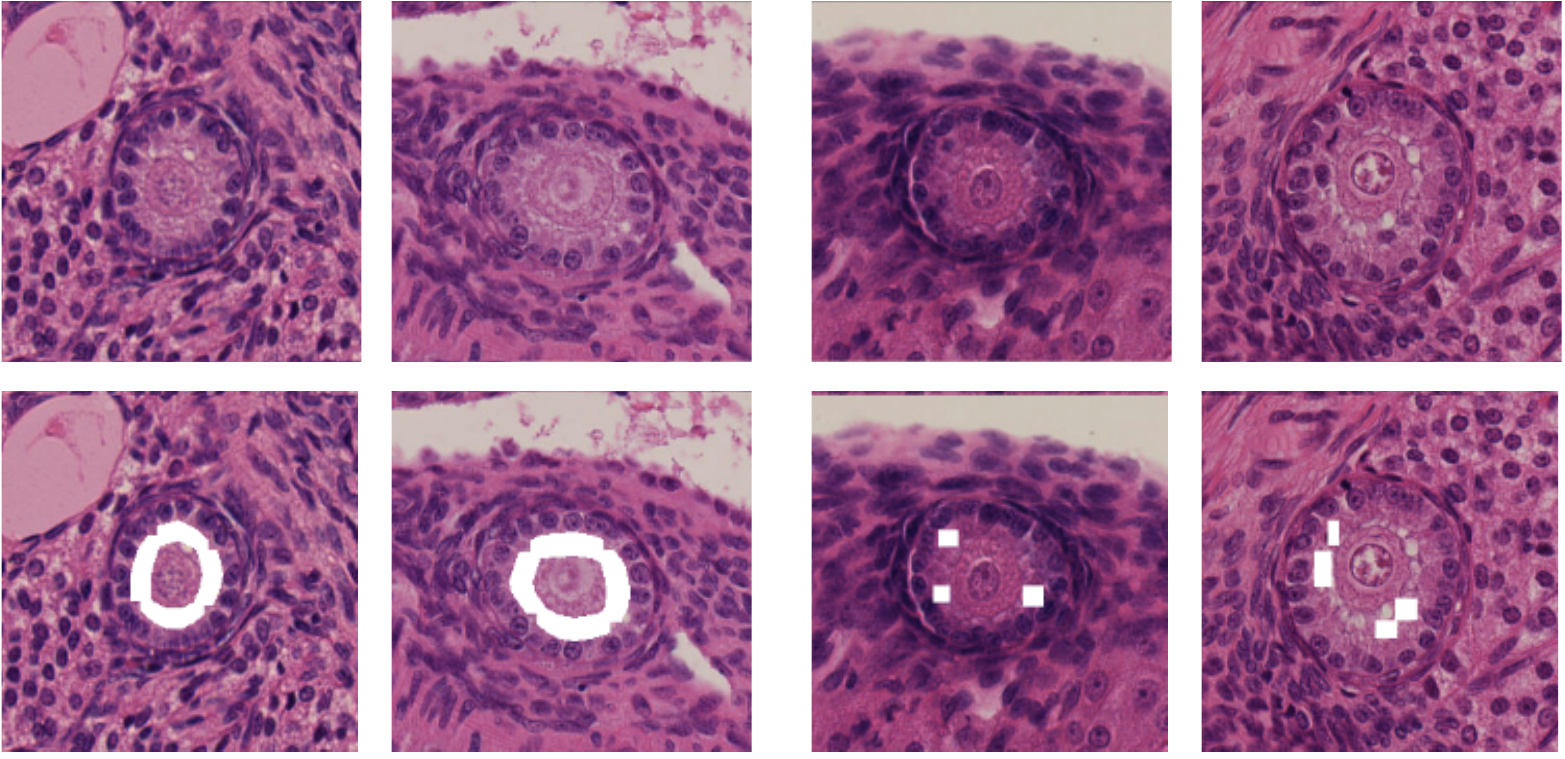}
  \caption{
  Columns 1 and 2 are \textit{Primordial follicles}.
  They have exactly one ring of Granulosa cells.
  Columns 3 and 4 are \textit{Growing follicles}.
  The have multiple rings of Granulosa cell.
  In order to distinguish them from Primordial Follicles one has to find a second layer of Granulosa cells.
  Features that should in theory support the classification decision are indicated by white rectangles.
  }
  \label{fig:follicle_mask_samples}
\end{figure}

We started our investigation with a trained Yolo model and a dataset of 1600 rats follicles, both provided by our industry partner. The trained Yolo model achieved high test set accuracy (cf.\ Figure \ref{fig:yolo_on_follicle_dataset} for a few examples). According to histopathology, a follicle can be recognized by having a spherical membrane surrounding it and a nucleus in the center. The distinction between primordial and growing follicles can be done by looking at the number of Granulosa cell layers (see Figure~\ref{fig:follicle_mask_samples}).

\begin{figure}[h]
  \centering
  \includegraphics[width=0.25\paperwidth]{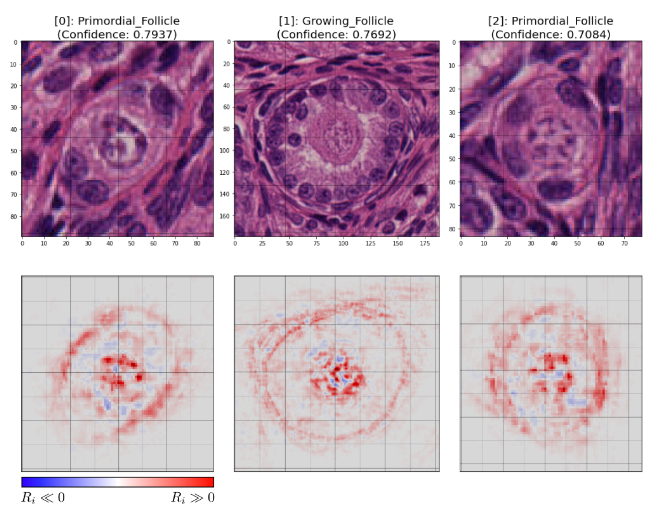}
  \caption{
  Heatmaps produced by LRP for the predictions made in Figure~\ref{fig:yolo_on_follicle_dataset}. Red color indicates pixels supporting the classification into the correct class.
  }
  \label{fig:lrp_on_yolo}
\end{figure}

To inspect the decision strategy used by our ML model, we adapted the LRP \cite{bach2015pixel} explanation technique to the Yolo architecture so that pixels supporting the Yolo model's decision are highlighted, resulting in heatmaps shown in Figure~\ref{fig:lrp_on_yolo}. We can observe that the expected strategy outlined above does not appear in LRP heatmaps. This indicates that the ML model uses a different strategy. However, the produced explanations fail to reveal what actual strategy the ML model uses. 

\begin{figure}[h]
  \centering
  \subfigure{\includegraphics[width=0.48\linewidth]{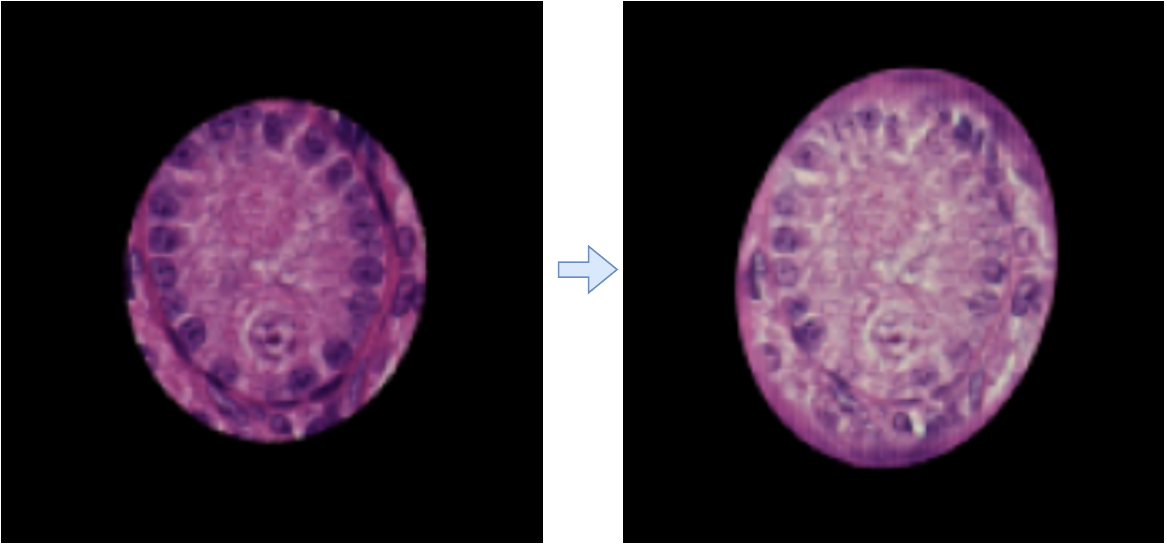}}
  \subfigure{\includegraphics[width=0.48\linewidth]{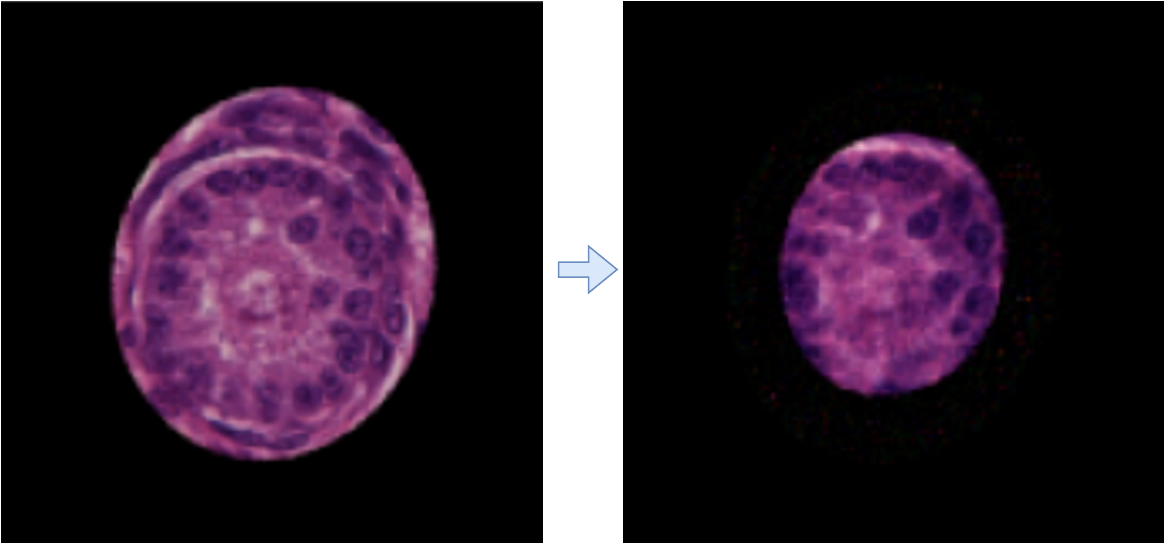}}
  \caption{
  \textit{Left:} produced counterfactuals transforming a primordial follicle into a growing follicle.
  \textit{Right:} counterfactuals for transforming a growing follicle into a primordial one.
  }
  \label{fig:confounder_size}
\end{figure}

Hence, we further explored the model's decision strategy using another type of XAI method: a \textit{counterfactual explainer}~\cite{dombrowski2021diffeomorphic, dombrowski2022diffeomorphic} (cf.\ Figure~\ref{fig:confounder_size} for the resulting explanations). A black mask is applied outside the incircle of the detected cells to ease the process of generating counterfactuals and a classifier is trained on the dataset that is created like this.
Interestingly, the new counterfactual explanations indicates a reliance of the classifier on the \textit{size} of the follicle, a confounder that was not apparent in previous explanations.
---This suggests that attribution-based explanations may only detect a subset of confounders and that refinement techniques based on counterfactuals are needed in complement to techniques based solely on classical attributions (such as presented in \cite{anders2022finding}).

\section{Methods}

In this section we first introduce a recently proposed Counterfactual Explanations method which generates plausible counterfactuals using a generative model.
Then we present our main contribution, counterfactual knowledge distillation (CFKD), which enables us to refine a model based on the counterfactuals and feedback of a teacher whether they are actually correct counterfactuals.
Finally, we present different teachers that can be used for CFKD and a novel metric called ``Feedback Accuracy'' to better predict the true generalization capabilities of the refined model.

\subsection{Diffeomorphic Counterfactuals with normalizing flows}

We base our counterfactual explainer on the work of Dombrowksi et al.~\cite{dombrowski2021diffeomorphic, dombrowski2022diffeomorphic} and give a short review of the method here.

Let $f \colon X \rightarrow \mathbb{R}^C$ be a classifier that assigns the probability $f(x)_c$ of input $x \in X$ belonging to class $c \in \{1, \dots, C\}$. Counterfactual explanations of $f$ provide minimal deformations $x' = x + \delta x$ such that the prediction of the classifier is changed.

\begin{figure}
  \centering
  \includegraphics[width=0.4\paperwidth]{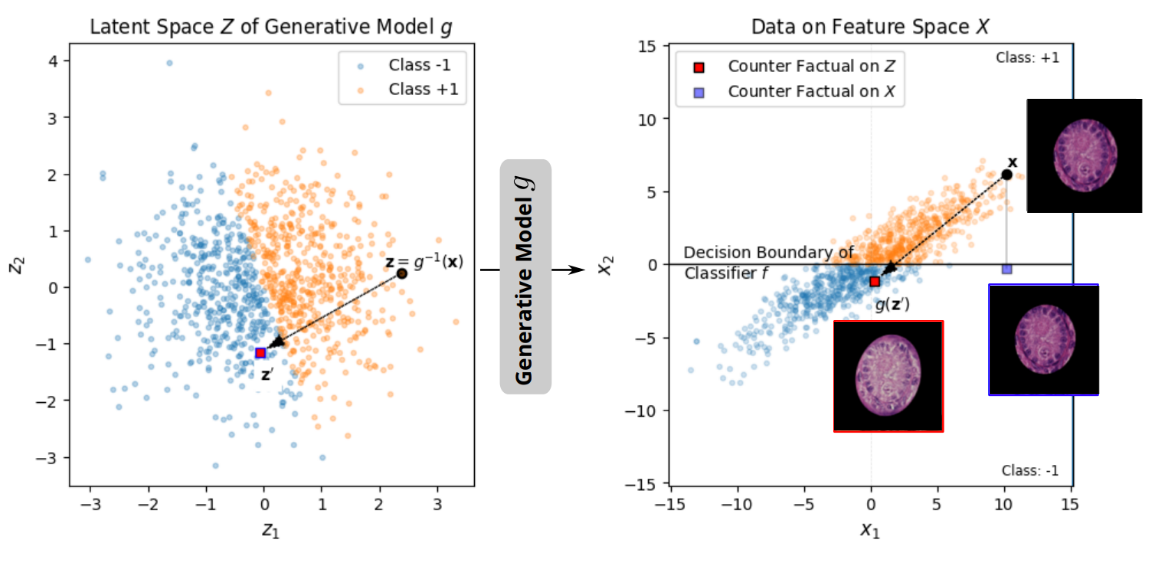}
  \caption{Diffeomorphic Counterfactual Explanations~\cite{dombrowski2022diffeomorphic}. As can be seen one has to use a generative model as a regularizer instead of directly using gradient ascend in input space to avoid creating adversarial examples.}
  \label{fig:cfs}
\end{figure}

Denote $c' \ne \arg\max_{c} f(x)_c$ be a target class. In practice, the perturbation $\delta x $ is  applied iteratively. Denote the perturbation at each step $\tau$ as $\delta x_\tau$ and $x_{\tau+1} = x_\tau + \delta x_\tau$. The update is repeated  until $\arg\max_c f(x_{\tau+1})_c =  c' $, and $x' = x_{\tau+1}$. A traditional approach is to use $\delta x_\tau $ proportional to $\partial f(x_\tau)_{c'} / \partial x_\tau$. But, this update step could lead to  changes in $x$ that is imperceptible, hence $x'$ being an adversarial sample. In many data modalities 
like images, the data lies approximately on a submanifold $D \subset X$ with significantly lower dimensionality $N_D$ than the dimensionality $N_X$ of the input space $X$. This is known as the manifold hypothesis. In such cases, it is more reasonable to find the counter factual $x'$ that is close to the data manifold $D$.   Denote $g: Z \rightarrow X$ be a generative model, mapping from a latent variable $z \in Z$ to a sample $x \in X$. ~\cite{dombrowski2021diffeomorphic, dombrowski2022diffeomorphic} show that optimizing on the latent space $Z$ of the generative model $g$ leads to  the update direction  $\delta x$ being close to the tangent space of the data manifold $D$ at the point $x_\tau$.  We refer to ~\cite{dombrowski2021diffeomorphic, dombrowski2022diffeomorphic} for the proofs. As a result, the procedure leads to meaningful changes in $x $ that are visible and semantically sound as illustrated in Figure~\ref{fig:cfs}.



Since our goal is not directly to create minimal counterfactuals, but to change the decision boundary of the student (by adding new samples directly on the decision boundary to the training dataset not much changes about the decision boundary) we do not stop directly when reaching the decision boundary, but empirically set the target confidence the counterfactual search should reach to $0.8$. Here a tradeoff between not entirely destroying the connection between the original sample and the counterfactual and creating very informative false counterfactuals deep on the other side of the decision boundary is to consider.

\begin{algorithm}[t]
\SetAlgoLined
\KwIn{Trained classifier $f$, dataset $D$, teacher $t$, number of iterations $n$}
\KwOut{Finetuned classifier $f'$}
\For{$i = 1$ to $n$}{
\For{$x, y \in D$}{
Choose $y_{target} \neq y$.
Generate counterfactual image $x^\prime$ based on $x$ and $y_{target}$;

$t(x, x^\prime)$ decides counterfactual is true counterfactual;

\If{True counterfactual}{
$y_{selected} = y_{target}$;
}
\Else{
$y_{selected} = y$;
}
Add the counterfactual image $x^\prime$ and $y_{selected}$ to the training data;
}
Retrain the classifier $f$ on the augmented training data;

Measure the feedback accuracy;
}
\KwRet{$f$}
\caption{Counterfactual Knowledge Distillation}
\label{alg:cfkd}
\end{algorithm}

\begin{figure}
  \centering
  \begin{subfigure}
    \includegraphics[width=\linewidth]{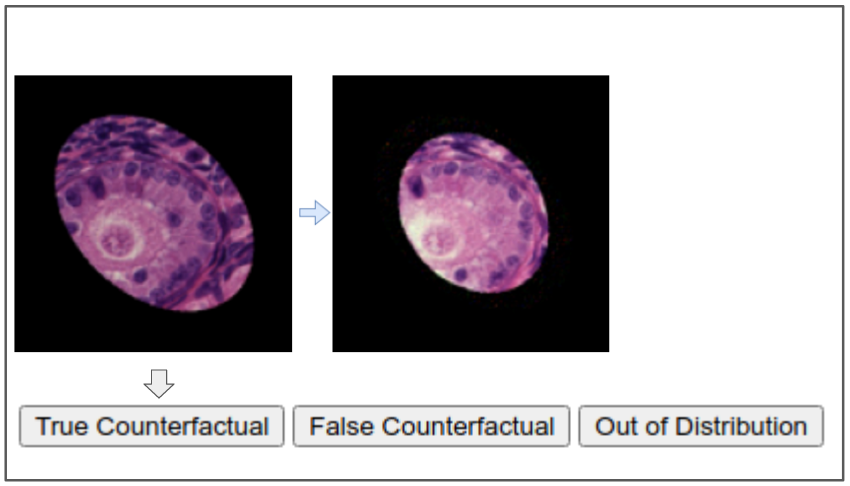}
  \end{subfigure}
  \hfill
  \begin{subfigure}
    \includegraphics[width=\linewidth]{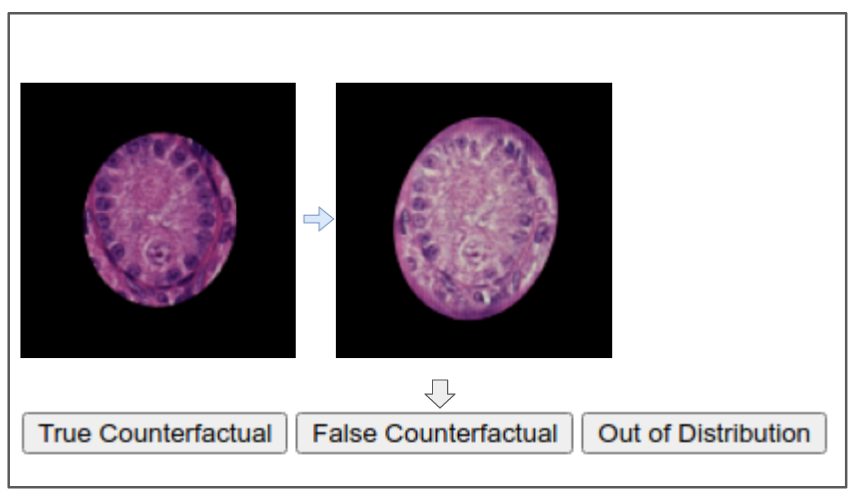}
  \end{subfigure}
  \caption{Comparison of two counterfactual cases.
  The top shows a \textbf{true counterfactual}, that agrees with intuitive decision boundary of human experts, since it actually changes from growing to primordial by removing the Granulosa cells.
  The bottom shows a \textbf{false counterfactual}, that is contrary to the intuitive decision boundary of human experts, because to change from a primordial to a growing follicle instead of adding Granulosa cells it increases the size instead.
  We embedded the feedback mechanism with the buttons in a web interface, that is started when executing CFKD.}
  \label{fig:virelay}
\end{figure}

\subsection{Counterfactual Knowledge Distillation}

The Counterfactual Knowledge Distillation algorithm (see Algorithm~\ref{alg:cfkd}) aims to distill knowledge from a teacher to a student and thereby improve the performance of it in the presence of confounding variables by iteratively generating counterfactual images, using them to determine whether a confounder is present, and then adding the counterfactual images to the training data with the appropriate label.
The algorithm needs a trained classifier, a dataset, and a teacher.
In each iteration, for each data point in the dataset, a counterfactual image is generated using an explainer.
The teacher determines whether the counterfactual image was created by changing the causal feature (True counterfactual) or a confounding feature (False counterfactual, see example in Figure~\ref{fig:virelay}).
If it is a true counterfactual, the target label of the counterfactual search is selected.
If not, the original label for the data point is selected.
The counterfactual image and selected label are then added to the training data.
The classifier is retrained on the augmented training data and the feedback accuracy (see section~\ref{sec:fa}) is measured.
The algorithm continues for a specified number of iterations and the refined model is selected based on the feedback accuracy.
Contrary to other knowledge distillation approaches the feedback can be given by a human since the (image, counterfactual)-pairs are human-interpretable.
Contrary to other confounder detection approaches feedback can be given by an oracle model, since the counterfactuals can be fed into the oracle model and assuming the oracle model is correct one can determine whether the correct feature was changed or a spurious one.
This is a very important benefit, since objective and resource-wise feasible experiments can be done with with the oracle model and for real applications where no oracle model exists human experts can be used.
CFKD has in principle the same robustness against teacher mislabels like supervised machine learning, which is trained with the same objective and also relies on human labels.

\subsection{Teachers}

There exist various alternatives that can serve as teachers.
The initial option tested, the oracle teacher, is rather theoretical in nature as it requires additional information regarding the underlying dataset.
Nonetheless, it facilitates conducting experiments on a large scale and produces easily reproducible results.
The other two teachers, namely, human and SpRAy, are universally applicable to any dataset provided that the user has expertise in determining the model's decision boundary.
Nevertheless, these options present the disadvantage of necessitating significant effort to conduct experiments on a large scale and producing results that are dependent on the quality of feedback provided by the human expert.
This paper proposes conducting experiments using the oracle teacher for the purpose of automating the workflow and producing reproducible outcomes.
However, we demonstrate that human feedback yields comparable results in one of the experiments. 

\subsubsection{Oracle model}

The Oracle model operates by utilizing an additional neural network that has been trained on a much more representative dataset, and is as a result not relying on confounders.
This neural network is employed to address the shortcomings of the student model by accepting/rejecting the counterfactuals produced by the latter.
This approach is particularly useful to evaluate the counterfactual technique by itself but is unsuitable in real-world setting because the extended dataset is not available (or if it were, the model should of course be trained on that extended dataset).

\subsubsection{Human-in-the-loop}

The human-in-the-loop teachers provide explanatory feedback through a human expert via a user interface.
We propose two possible human-in-the-loop teachers.
The human teacher (as illustrated in Figure~\ref{fig:virelay}) is a simple web interface for sequentially evaluating pairs consisting of an image and a counterfactual, and making a determination as to whether the explanation provided is accurate or not.
While this approach offers a high level of precision, it is also time-consuming and resource-intensive, particularly if an expert with significant domain knowledge is required.
The SpRAy teacher is designed to reduce the effort required by human experts to provide feedback and is based on the work of Anders et al.~\cite{lapuschkin2019unmasking} and implemented with Virelay~\cite{anders2021software}.
It accomplishes this by utilizing t-SNE~\cite{van2008visualizing} to embed the differences between the latent representation of an image and its corresponding counterfactual into a two-dimensional space.
The latent representation is directly extractable by the generative model utilized for building the counterfactuals.
The counterfactuals that are generated using similar strategies are then clustered together in the embedded space.
The human expert can then identify and select clusters of invalid strategies by drawing boxes around them.
These clusters of invalid data points are returned as negative feedback, while the rest are considered positive.
This approach is less labor-intensive than the human teacher, as it enables the human expert to review multiple strategies simultaneously, reducing the need to evaluate each sample individually.
Additionally, the clustered display of similar strategies can help the expert more easily identify the actual strategy being used.
However, this method relies on the assumption that t-SNE can accurately cluster the strategies, which may be less precise if a broad range of incorrect strategies are employed or the latent space of the generative model is highly entangled.



\subsection{Feedback Accuracy}
\label{sec:fa}

When the training and validation datasets are derived from the same pool of data the resulting model's validation accuracy cannot be trusted as a reliable indicator of its true performance. However, we still require a performance indicator to estimate the model's true test performance.
To address this issue, we introduce the feedback accuracy, defined as follows:
%
    $\mathrm{ACC}_\mathrm{feedback} = \frac{N_\mathrm{correct}}{N_\mathrm{total}}$.
%
This accuracy measures the proportion of counterfactuals generated on the validation dataset that are accepted as legitimate by the teacher.
In essence, $\mathrm{ACC}_\mathrm{feedback}$ tracks the teacher's evaluation of the model's ability to generate counterfactuals that correctly identify the generalizing strategy, which is assumed to be the solution to the real task.

The intuition behind the name counterfactual knowledge distillation is that if one would execute Algorithm~\ref{alg:cfkd} longer and longer the decision strategy of the student will become more and more like the teachers.
Furthermore, the feedback accuracy will go against one since both the student and the teacher would decide more and more similar.
Hence, the teacher distills its knowledge into the student based on the counterfactual explanations.

\section{Related Work}


\textbf{Clever Hans Detection}
Reducing reliance on confounders (Clever Hans effects) is a two step process, which first requires to \emph{find} such effects before they can be removed.
Methods such as Spectral Signature \cite{tran2018spectral} are able to identify backdoors in neural networks, which are a particularily \emph{strong} type of confounders, as they must be the strongest feature in order surpass the true class of the sample.
Spurious correlations which lead to Clever Hans effects are \emph{weak} confounders that influence predictions less, since they are always found alongside true class-relevant features.
Apart from manual inspection of the samples, feature attribution approaches \cite{bach2015pixel} can provide more insight into whether features are relevant for a model's prediction.
Explanatory Interactive Learning (XIL) \cite{schramowski2020making,teso2019explanatory} finds weak confounders by presenting features attributions to human experts.
Spectral Relevance Analysis (SpRAy) \cite{lapuschkin2019unmasking,anders2022finding} attempts to find confounders by presenting \emph{embeddings} and \emph{clusterings} of feature attributions to human experts in order to reduce the work load.
It is arguably hard for an expert to find confounders in plain samples or feature attributions as provided by XIL, or even in clusters and embeddings as provided by SpRAy.
Our approach simplifies the task presented to the expert, as they only need to decide whether a counterfactual has or hasn't shifted away from the actual class.

\textbf{Feature Unlearning} Various approaches for the alleviation or removal of confounders have been developed beyond the basic removal of afflicted samples from the training data.
While some methods provide their own or derived approaches for confounder detection \cite{schramowski2020making,teso2019explanatory,anders2022finding}, other works assume the confounders are known a priori.
For instance, eXplanatory Interactive Learning (XIL) uses human generated labels during training to provide feedback to the model by replicating samples affected by CH phenomena and replacing the confounding features with noise or otherwisely generated patterns~\cite{schramowski2020making, teso2019explanatory}.
The work of Kim et al.~\cite{kim2019learning} introduces a model regularization scheme, in which an additional ``artifact detector'' learning specific biasing features is attached to the original predictor.
The original model is then driven to minimize the shared information with the dedicated bias predictor, and thus to unlearn to use artifactual features for inference.
Ross et al.~\cite{ross2017right} aim to guide the model towards the correct behavior by penalizing high attribution scores in undesired regions by extending the optimization function with a ``Right for the Right Reasons (RRR)'' loss term.
Similarly, Rieger et al. \cite{rieger2020interpretations} propose Contextual Decomposition Explanation Penalization (CDEP), a method for regularizing model behavior based on explanations obtained from Contextual Decomposition (CD)~\cite{ murdoch2018beyond}, by complementing the classification error of the loss function with an explanation error term.

\textbf{Class Artifact Compensation}
To the best of our knowledge, the approach we propose is most closely related to Class Artifact Compensation (ClarC)~\cite{anders2022finding}.
In the original work, the confounder detection is implemented through an extended Spectral Relevance Analysis, yielding class-wise labels for each confounder. 
Feature unlearning is accomplished using these confounder labels through two variants of the method: augmentative (A-ClarC) and projective (P-ClarC).

The feature unlearning provided through ClArC can be decoupled from the confounder detection provided by SpRAy.
To detect confounders, our work presents counterfactuals to a human expert, while SpRAy presents embeddings and clusterings of feature attributions.
In ClArC, the confounder labels are used to construct a confounder model (either explicit or using a linear model motivated by Concept Activation Vectors (CAV) \cite{kim2018interpretability}) to either introduce the confounder into unafflicted samples for augmented fine-tuning in A-ClArC, or to remove CAV of the confounder through linear projection in P-ClArC. 


Our proposed method, Counterfactual Knowledge Distillation (CFKD), can be considered as a confounder detection approach coupled with a version of A-ClarC with the added benefit that our counterfactual generation already provides a confounder model in order to remove or introduce confounders used for the augmented fine-tuning.

\section{Experiments}

This section presents four experiments that demonstrate the capabilities of the proposed CFKD approach:
1) A qualitative analysis shows the ability of CFKD to adjust the decision strategy of a follicle classifier to align with the biologically correct strategy.
2) An extensive experiment with different augmentation-based confounders and poisoning levels on CelebA is conducted to evaluate the performance of counterfactual knowledge distillation in addressing the problem with confounders in a BlondHair classifier. The results demonstrate the effectiveness of the proposed approach.
3) The ability of counterfactual explanations to identify and quantitatively address confounders in subsampled real-world datasets is demonstrated using the colorectal cancer tissue 100K dataset~\cite{https://doi.org/10.5281/zenodo.1214456} with a known staining issue. CFKD is shown to be effective in addressing these confounding factors as well.
For all our experiments we used ResNet-18~\cite{he2016deep} as classifier and Glow~\cite{kingma2018glow} as generative model.

\subsection{Qualitative Results on the Follicle Dataset}

\begin{figure}[h]
  \centering
  \subfigure{\includegraphics[width=0.48\linewidth]{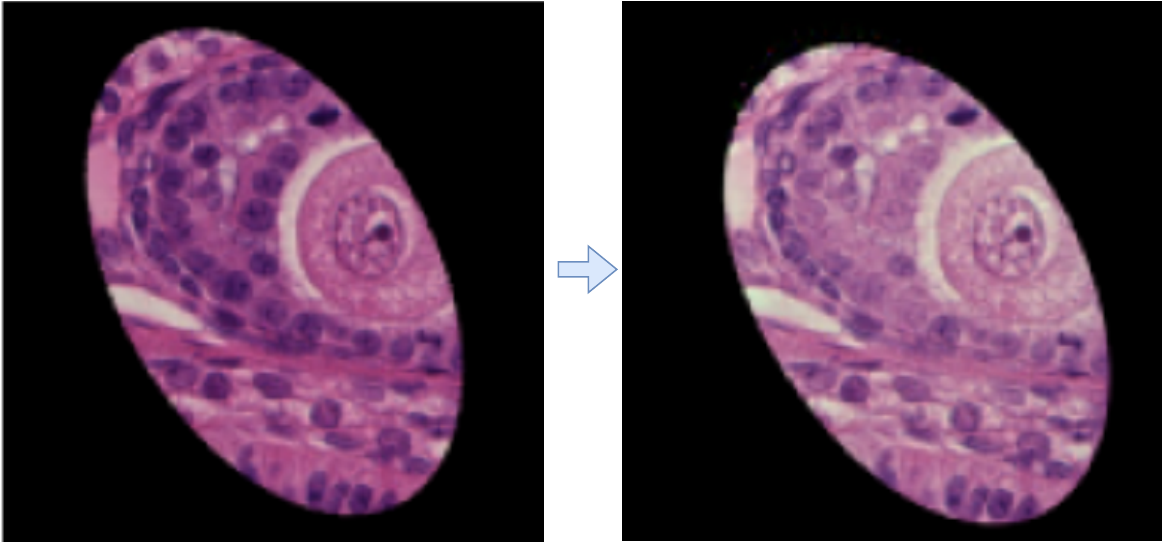}}
  \subfigure{\includegraphics[width=0.48\linewidth]{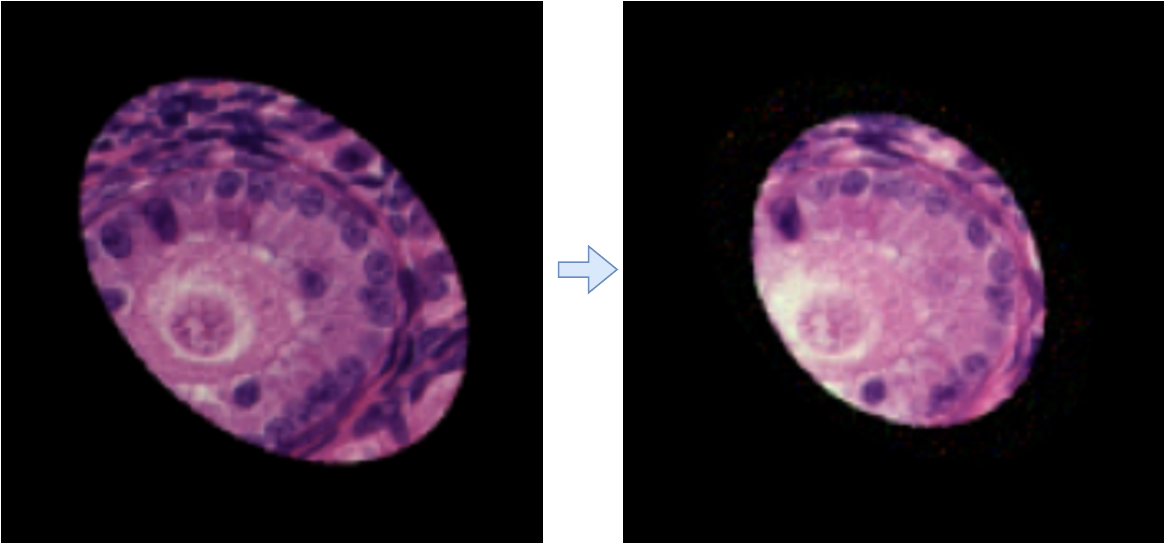}}
  \caption{
  Counterfactual explanations after applying CFKD on the follicle dataset, and exhibiting a change of strategy after teaching.
  (Compare to Figure\ \ref{fig:confounder_size}). 
  }
  \label{fig:follicles_qualitative}
\end{figure}

We apply our method to the motivating example of Section \ref{section:motivation}. Note that there is no unpoisoned dataset and no reproducible teacher model for this showcase (the only one available is our own interaction via the human-in-the-loop teacher), hence, we report qualitative results and defer quantitative evaluation to the next experiment. We performed 3 iterations of CFKD with 100 counterfactuals each.
Counterfactuals produced after teaching are given in Figure\ \ref{fig:follicles_qualitative} (compare with Figure\ \ref{fig:confounder_size} before teaching).

One can see, that instead of just changing the size of the follicle now the counterfactual explainer attempts to remove the Granulosa cells in order to transform growing follicles to primordial follicles, thereby showing that the decision strategy lost much of its `Clever Hans' nature and now relies on the more meaningful Granulosa cells patterns.

\subsection{CelebA Dataset with augmentation-based Confounders}
\label{sec:celeba}

\begin{figure*}
  \centering
  \subfigure{\includegraphics[width=0.3\linewidth]{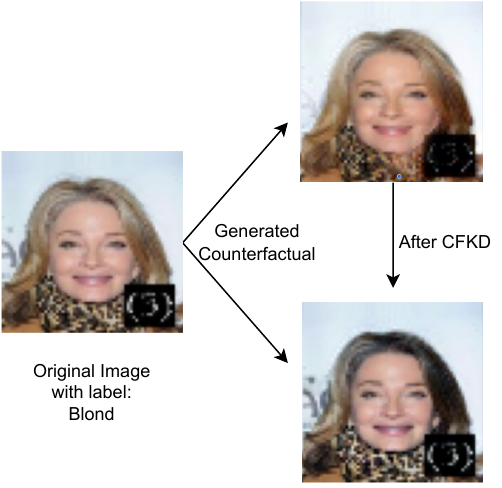}}
  \subfigure{\includegraphics[width=0.3\linewidth]{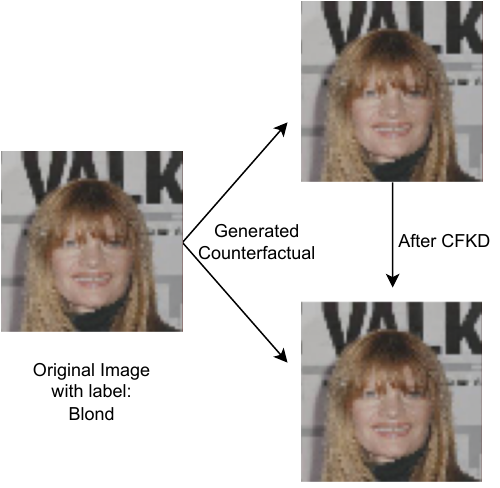}}
  \subfigure{\includegraphics[width=0.3\linewidth]{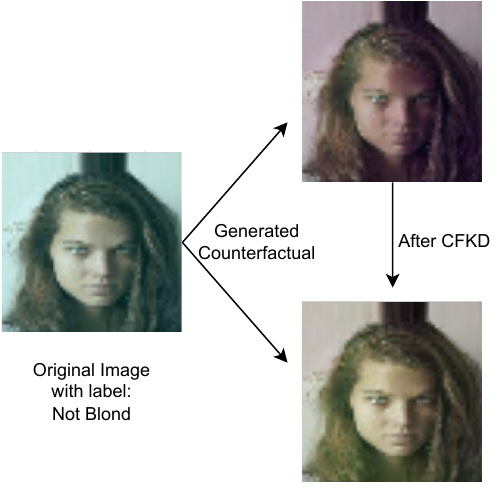}}
  \caption{Qualitative result for CelebA with 100\% poisoning. The leftmost sample come from the copyright tag poisoned dataset, the middle sample from the intensity shifted dataset and the rightmost from the color shifted dataset. One can see that before applying CFKD the counterfactuals are created by changing the confounder e.g. on the left by blurring the copyright tag. After CFKD the true feature, namely the hair color is changed. However, in some case like the middle sample it is not that obvious what was changed by the model and what changed after applying CFKD even though we know from Figure~\ref{fig:quantitative_celeba} that CFKD quantitatively works.}
  \label{fig:qualitative_celeba}
\end{figure*}

\begin{figure*}
  \centering
  \subfigure{\includegraphics[width=0.3\linewidth]{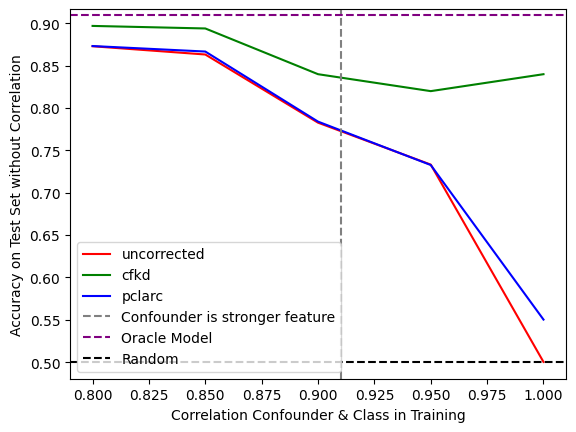}}
  \subfigure{\includegraphics[width=0.3\linewidth]{figures/intensity_quantitativ.pdf}}
  \subfigure{\includegraphics[width=0.3\linewidth]{figures/color_quantitativ.pdf}}
  \caption{
  Quantitative Results on CelebA.
  One can see, that with stronger correlation between the BlondHair class and the confounders the unpoisoned test accuracy of the uncorrected classifier drops drastically.
  For all modalities CFKD systematically improves the results significantly.
  }
  \label{fig:quantitative_celeba}
\end{figure*}

\begin{figure*}
  \centering
  \subfigure{\includegraphics[width=0.33\linewidth]{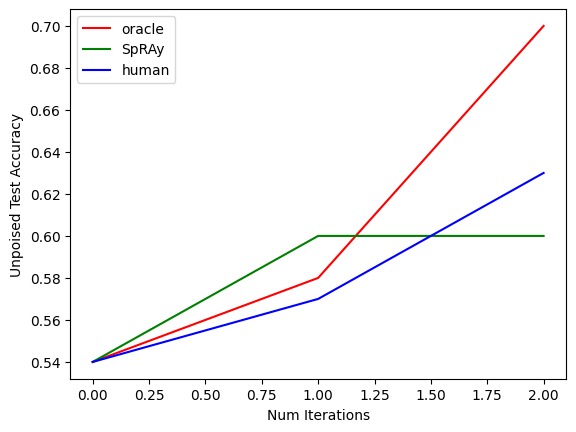}}
  \subfigure{\includegraphics[width=0.33\linewidth]{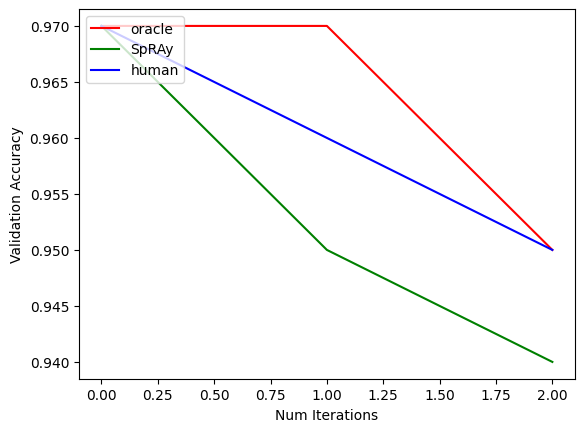}}
  \subfigure{\includegraphics[width=0.33\linewidth]{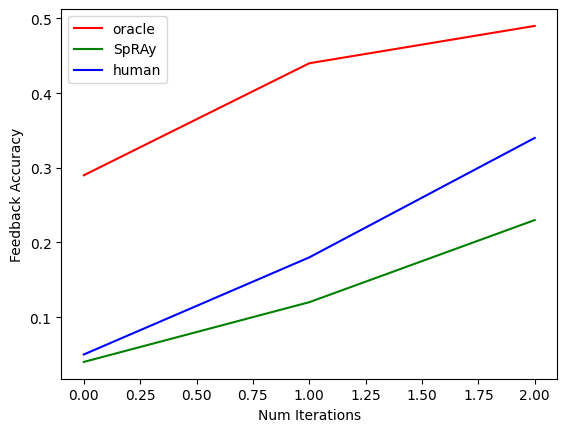}}
  \caption{
  Training ablation in regard to different trainers for a 100\% poisoned CelebA with intensity shift.
  Shows that validation accuracy which is higher than the oracle test performance in the beginning is anti proportional to unpoisoned test accuracy while feedback accuracy is proportional.
  Furthermore, it shows that Oracle, Human and SpRAy teacher behave similar.
  }
  \label{fig:celeba_training}
\end{figure*}

As a second experiment, we consider a subset of CelebA, where we select 40K samples, so that 50\% of the depicted persons are blond and 50\% are not.
Furthermore, we created 3 versions of such a dataset each of them poisoned with a particular type of confounder: one with a copyright tag in the bottom right of the image that mimics known confounders like the copyright tag on PASCAL VOC, one with an intensity shift of the images and one with a color shift.
In order to mimic our motivational example more closely we smooth these confounders and base them on a continuous transition.
Concretely, the copyright tag confounder is based on a transition from being 100\% opaque to being 0\% opaque and a sample is considered having a confounder when the opaqueness is lower than 50\%.
The intensity shift confounder is based on adding up to 24 (on the 0--255 intensity scale) on each pixel of the samples with the confounder and subtracting up to 24 of each pixel for the samples without confounder and then clipping all pixel values back to a range between 0 and 255.
The color shift is created by adding up to 24 on the red channel and subtract up to 12 of the green and the blue channel of the image for the images with confounder and vice verse for the ones without.
For each version 36K images are used for training and validation and 4K samples are held out as test set.
Therefore we ensure, that the percentage of all four possible combinations of containing a person with blond hair and having the confounder are 25\% both in the training and the test set.
For our poisoned classifiers we select the 16K sample subset of the dataset, that only contains images, that contain a proportion of $\alpha$ images that either contain a blond person and the confounder or that contain no blond person and no confounder and $1-\alpha$ images that either contain a blond person without containing the confounder or a non-blond person that contains the confounder.
However, we still ensure that this subset contains the same amount of blond as non-blond persons.
We refer to these datasets as the poisoned datasets.
Training classifiers on these achieves close to perfect accuracy on their training and validation set.
However, if the classifier is evaluated on the unpoisoned test dataset, where there is no correlation between having a confounder and the person being blond they achieve down to 50\% unpoisoned test accuracy (see Figure~\ref{fig:quantitative_celeba}), which is the random performance due to the balanced nature of the dataset.

Further analysis shows that most errors are made on pictures that either contain a blond person but no confounder or contain no blond person, but a confounder.
When training a generative model on the poisoned dataset and creating a counterfactual explanation we can see in Figure~\ref{fig:qualitative_celeba} on the top left, that the counterfactual for a blond person is created by removing the copyright tag and not by changing the color of the hair.
Next we train an oracle classifier, that objectively mimics human feedback on the full training dataset.
Probably due to dataset noise (e.g. some people in CelebA are bold or wear a hat) it is also not able to achieve 100\% accuracy on the test set, but at least it achieves 91\%.
Now we use this oracle classifier to give feedback for CFKD.
We used CFKD with 500 randomly sampled datapoints from the dataset per iteration for which counterfactuals are build and a maximum of 5 iteration except for the strongest poisoning where the maximum are 12 iterations.
We can see in the bottom of Figure~\ref{fig:qualitative_celeba}, that the counterfactual explanation for the same sample changes towards actually changing the color of the hair and not removing the confounder anymore.
Furthermore, as can be seen in Figure~\ref{fig:quantitative_celeba}, the unpoisoned test accuracy is getting closer to the classification accuracy of the oracle classifier.
In Figure~\ref{fig:celeba_training} we show, that the feedback accuracy is correlated to the unpoisoned test accuracy while the validation accuracy is not.
Moreover, we show that all suggested teachers behave similar.

Due to the possible interpretation of CFKD as A-ClarC we compared with P-ClarC (since A-ClarC does not provide an automated way how to augment samples) and also provide our implementation of P-ClarC.
However, P-ClarC was developed for a setting where the confounder is only present on one class and is a localized object that is either there or absent.
Furthermore, it assumes that at least one sample of the class does not contain the confounder.
Our experiment setting derived in analogy to our motivational example potentially has confounders in both classes.
Moreover, our definition of a confounder is based on the threshold of a continuous transition from an underlying variable like the size of the follicle, the opaqueness of the copyright tag, the strength of the intensity/color shift.
In the case of the strongest poisoning of 100\% there is not a single sample of a blond person without a confounder.
Additionally, deriving feedback can not be automated with an oracle model and for distributed confounders like the intensity / color shift it is not possible for a human expert to see them based on the attribution maps, since they mark the whole image.
Even for the copyright tag confounder an attribution map that highlights the region where the copyright tag is can not be guaranteed to detect an artifact, since it is not clear whether the region of the image was marked because there is a copyright tag with low opaqueness or since it was solely marked due to the model measuring the opaqueness of the copyright tag and the opaqueness might in fact be high.
Due to this reasons our implementation of P-ClarC was not capable to improve over the uncorrected classifier at all.

In a summary CFKD is very effective in removing confounders qualitatively and quantitatively and we not aware of related work that could do it in the same way.

\subsection{Colorectal Cancer Tissue Dataset}

\begin{figure}[h]
  \centering
  \subfigure{\includegraphics[width=0.48\linewidth]{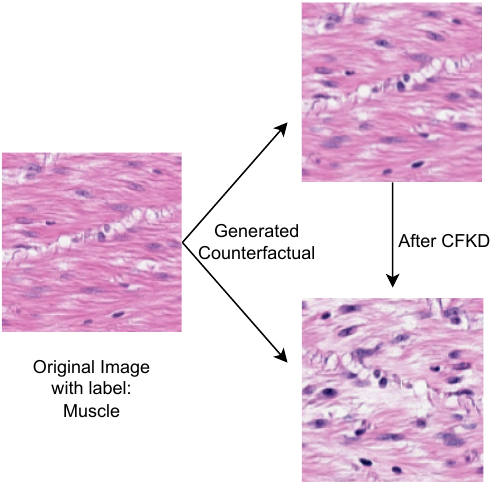}}
  \subfigure{\includegraphics[width=0.48\linewidth]{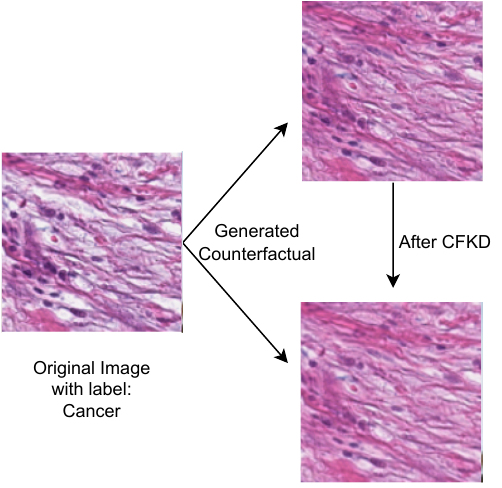}}
  \caption{
  The figure shows the qualitative results of the correction of the confounder on the Colorectal cancer tissue dataset.
  }
  \label{fig:cancer_tissue_samples}
\end{figure}

\begin{figure}[h]
  \centering
  \includegraphics[width=0.2\paperwidth]{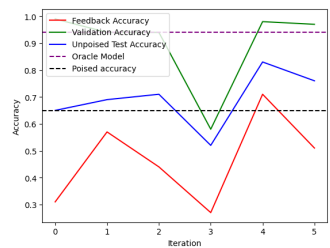}
  \caption{
  One can see, that CFKD effectively increases the unpoisoned test accuracy in the quantitative experiment on the colorectal cancer tissue datset.
  Furthermore, it shows that the peak in feedback accuracy coincides with the peak in unpoisoned test accuracy while the peak in validation accuracy does not.
  }
  \label{fig:cancer_tissue_quantitative_results}
\end{figure}

In this experiment, we investigate the performance of a classifier trained on the colorectal cancer tissue 100K dataset~\cite{https://doi.org/10.5281/zenodo.1214456}, which includes nine classes of histopathological images.
We address the issue of staining-based confounders, which is well-known in the histopathology community, especially when distinguishing between benign muscular tissue and malign cancer-associated stromas.
To address this issue, we train a classifier that only separates these two classes. 
To create an oracle classifier, we utilize the full dataset.
Then subsample a poisoned training dataset where the strength of the hematoxylin staining is strongly correlated with the cancer-associated stroma and train our uncorrected model on it.
Our explainer effectively utilizes this correlation to create counterfactuals, as demonstrated in Figure~\ref{fig:cancer_tissue_samples} (top row).
Whether the change in strategy caused by CFKD and shown in Figure~\ref{fig:cancer_tissue_samples} is reasonable from a medical viewpoint remains open for interpretation, but the CFKD corrected model performs quantitatively better on the test dataset without the correlation between the hematoxylin stain and the class, as illustrated in Figure~\ref{fig:cancer_tissue_quantitative_results}.
Overall, our experimental results highlight the effectiveness of our approach in improving the classifier's decision strategy and subsequent performance on the challenging colorectal cancer tissue 100K dataset.

\section{Conclusion}

In conclusion, we have presented a novel technique, CFKD, to detect and remove reliance on confounders in deep learning models by generating counterfactuals and leveraging explanatory feedback.
We demonstrated the effectiveness of CFKD on both augmented and real-world datasets, and could measure the performance increase either on the true test accuracy when available or using our newly proposed feedback accuracy metric.
Our method is flexible, and can work together with a variety of teacher models and explanation/counterfactual generation techniques.
Overall, the paper contributes to improving the interpretability and reliability of deep learning models, and has the potential to impact a wide range of applications.


\section{Acknowledgements}

This work was supported by the German Ministry for Education and Research (BMBF) under Grant 
01IS18037A, 
and by BASLEARN---TU Berlin/BASF Joint Laboratory, co-financed by TU Berlin and BASF SE. P.C.\ is supported by the German Federal Ministry of Education and Research and the Max Planck Society.

{\small
\bibliographystyle{ieee_fullname}
\bibliography{egbib}
}

\end{document}